\documentclass[journal]{IEEEtran}
\usepackage[caption=false,font=normalsize,labelfont=sf,textfont=sf]{subfig}
\usepackage[pdftex]{graphicx}
\usepackage{stfloats}
\usepackage{rotating}
\usepackage{graphicx}  
\usepackage{epstopdf}
\usepackage{amsmath}
\usepackage{booktabs}
\usepackage{amssymb}
\usepackage{textcomp,booktabs}
\usepackage{multirow}
\usepackage[usenames]{color}
\usepackage{wrapfig}
\usepackage{longtable}
\usepackage{colortbl,booktabs}
\usepackage{verbatim}

\begin{document}
%

\title{A Novel Video Salient Object Detection Method via Semi-supervised Motion Quality Perception}




\author{Chenglizhao Chen$^{1}$\thanks{Corresponding
		author: Chong Peng (pchong1991@163.com)}
	~~~~Jia Song$^{1}$~~~~Chong Peng$^{1*}$~~~~Guodong Wang$^{1}$~~~~Yuming Fang$^{2}$\\
$^1$Qingdao University~~~~~$^2$Jiangxi University of Finance and Economics\\
}

\markboth{IEEE Transactions on Image Processing, VOL.XX, NO.XX, XXX.XXXX}%
{Shell \MakeLowercase{\textit{et al.}}: Bare Demo of IEEEtran.cls for Journals}

\maketitle

\begin{abstract}

Previous video salient object detection (VSOD) approaches have mainly focused on designing fancy networks to achieve their performance improvements. However, with the slow-down in development of deep learning techniques recently, it may become more and more difficult to anticipate another breakthrough via fancy networks solely.
To this end, this paper proposes a universal learning scheme to get a further 3\% performance improvement for all state-of-the-art (SOTA) methods.
The major highlight of our method is that we resort the ``motion quality''---a brand new concept, to select a sub-group of video frames from the original testing set to construct a new training set.
The selected frames in this new training set should all contain high-quality motions, in which the salient objects will have large probability to be successfully detected by the ``target SOTA method''---the one we want to improve.
Consequently, we can achieve a significant performance improvement by using this new training set to start a new round of network training.
During this new round training, the VSOD results of the target SOTA method will be applied as the pseudo training objectives.
Our novel learning scheme is simple yet effective, and its semi-supervised methodology may have large potential to inspire the VSOD community in the future.

\end{abstract}

\maketitle
\IEEEpeerreviewmaketitle

\section{Introduction and Motivation}
Different from images that comprise spatial information only, video data usually contain both spatial (appearances) and temporal (motions) information.
To alleviate the computational burden, most of the video related applications~\cite{yan20203d,yan2020deep,belloulata2014object,chen2015real,fan2019metrics,chen2016robust,peng2020robust} have adopted the video salient object detection (VSOD) approaches as the pre-processing tool to filter the less important video contents while highlighting the salient objects that attract our visual system most, aiming to strike the trade-off between efficiency and performance.

\begin{figure*}[t]
	\centering
	\includegraphics[width=0.9\linewidth]{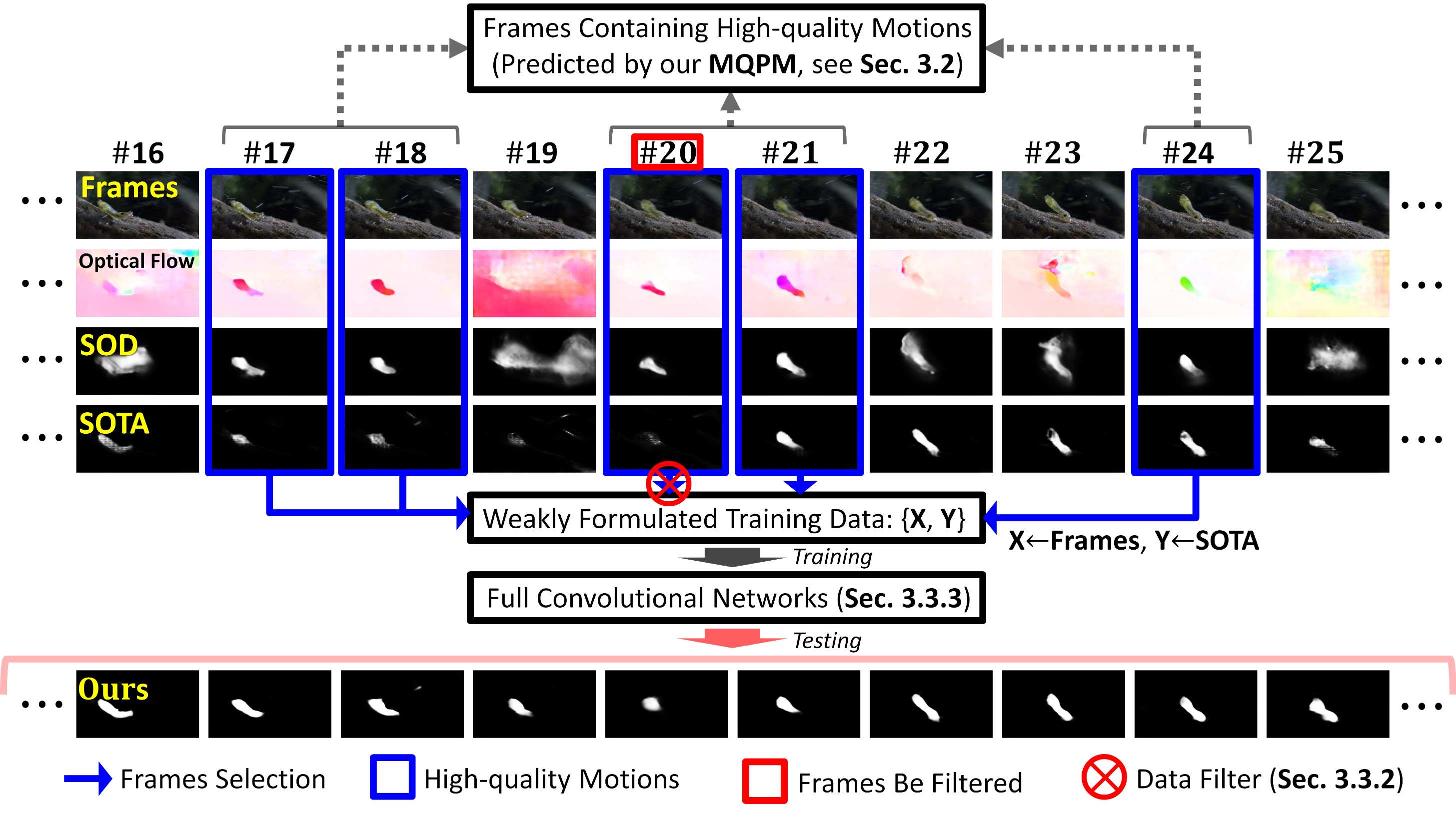}
    \vspace{-0.6cm}
	\caption{The key motivation of our method is to select a sub group video frames from the original testing set to construct a new training set, and these selected frames should have high-quality motions (by our MQPM) and their VSOD provided by the target SOTA method will be used as the pseudo-GT to start a new round training which will improve the target SOTA method significantly.
		\textbf{SOD}: the salient object detection results by feeding the optical flow data into the pre-trained image salient object detection model (we choose CPD~\cite{wu2019cascaded_cpd} here, see Eq.~\ref{eq:MQS}); \textbf{SOTA}: the VSOD of the target SOTA method (we take the SSAV~\cite{fan2019shifting_ssav} for example) which we aim to improve its performance, and it can be any other SOTA methods; \textbf{Ours}: the final VSOD results after using our novel learning scheme, of which the overall performance have outperformed the SOTA results, significantly.}
    \vspace{-0.2cm}
	\label{fig:motivation}
\end{figure*}

After entering the deep learning era, the state-of-the-art (SOTA) VSOD approaches have achieved steady performance improvements via various fancy networks, such as ConvLSTM~\cite{xingjian2015convolutional} and 3D ConvNet~\cite{tran2015learning_3d}.
However, with the slow-down in development of the deep learning techniques recently, we shouldn't anticipate for new breakthrough via fancy networks solely.
For example, compared with the leading SOTA method in 2019 (i.e., MGA~\cite{li2019motion_mga}), the performance improvement made by the most recent work in 2020 (i.e. PCSA~\cite{gupyramid_pcsa}) is really marginal with a performance gap less than \underline{1}\% averagely.
This fact motivates us to wonder why wouldn't we develop a universal learning scheme, rather than using fancy networks, to get the SOTA performances further improved?

Given an off-the-shelf VSOD approach (we name it as the ``target SOTA method''), this paper aims to improve its performance via a novel learning scheme, and we formulate our idea as following.\\
\underline{\textbf{1)}} We shall select a sub-group of video frames from the original testing set to construct a new training set, and these selected frames are needed to be the ones that have been ``successfully detected'' by the target SOTA method.\\
\underline{\textbf{2)}} Consequently, we will achieve a significant performance improvement by using this new training set to start a new round of network training, in which the VSODs of the corresponding SOTA method will be used as the training objectives (pseudo-GT).\\
So, without using any saliency ground truth (GT) of the original testing set, all that remains now is how can we know which frames will be successful detected by the target SOTA method in advance.

Our key idea is quite simple and straight-froward, which is inspired by a common phenomenon in the SOTA methods; i.e., for most of the SOTA VSOD methods, their performances usually vary from frame to frame, even though these frames belong to an identical video sequence sharing similar scenes.
For example, as is shown in Fig.~\ref{fig:motivation}, the 1st row shows 10 consecutive frames with similar scenes containing a \emph{worm} as the salient object; however, as is shown in the 3rd row, the VSOD results of the SOTA method (SSAV~\cite{fan2019shifting_ssav}) in the frame \#17, \#18, \#21 and \#24 are clearly better than other frames.
The main reason is that the VSOD performance is determined by both spatial and temporal saliency clues.
Though the spatial saliency clues are usually stable between consecutive video frames, the motion saliency clues may vary a lot due to the unpredictable nature of movements, not to mention other additional challenges induced by camera view angle changes.
So, we propose a brand new concept---``motion quality'', to predict which video frames will have large probability to be successfully detected by the target SOTA approach.

\begin{figure}[!h]
	\centering
	\includegraphics[width=0.9\linewidth]{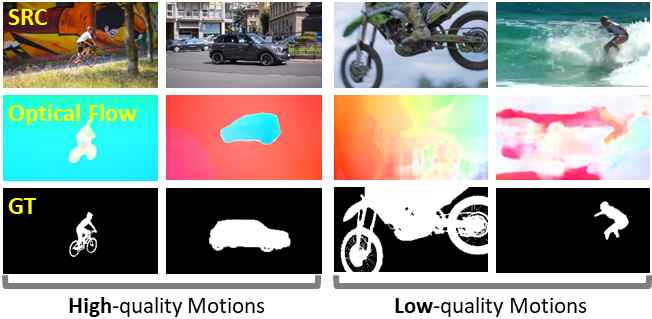}
    \vspace{-0.2cm}
	\caption{Motion quality demonstrations, where the high-quality motions can usually separate salient objects from their non-salient surroundings nearby, while the low-quality motions cannot achieve this.}
    \vspace{-0.2cm}
	\label{fig:MotionQDemo}
\end{figure}

For those clear motions (e.g., rigid movements) which can positively facilitate the VSOD task by separating salient objects from their non-salient surroundings nearby, we name it as the ``high-quality motions'', and we call other cases as the ``low-quality motions'' accordingly, see Fig.~\ref{fig:MotionQDemo}.

In most cases, we believe that those video frames containing ``high-quality motions'' should be selected into our new training set.
To predict motion quality in advance, we advocate a semi-supervised scheme to train our motion quality perception module (MQPM) within a frame-wise manner, see the Fig.~\ref{fig:overview}-C and it will be detailed in Sec.~\ref{sec:MQPM}.
As one of the key components in our learning framework, the MQPM takes motion patterns (sensed by optical flow) as input, and then it makes a binary decision regarding whether or not the given frame contains high-quality motions.
Meanwhile, in the case of a video frame has some high-quality motions, the MQPM will also provide the corresponding spatial locations of these high-quality motions, and these spatial locations will be used to facilitate the data filtering scheme (Sec.~\ref{sec:DF}), another key component in our learning framework, to double-check if these motions really belong to the high-quality cases.

In summary, the main contributions of our method can be summarized as following four aspects:
\begin{itemize}
	\item
	A semi-supervised learning scheme to conduct Motion Quality Perception (to the best of our knowledge, this is the first attempt to improve the VSOD performance from the motion quality perspective);
	\item
	A universal scheme to improve the performance of any other SOTA methods (at least 3\% performance improvement in general);
	\item
	Extensive quantitative validations and comparisons (almost all SOTA methods in recent 3 years over 5 largest datasets);
    \item Method source code and results are all publicly available at       \underline{\textbf{{https://github.com/qduOliver/MQP}}}, which will have large potential to benefit the VSOD community in the future.
\end{itemize}

\section{Related Work}

\subsection{Image Salient Object Detection}
The main target of image saliency (\cite{han2018advanced,han2017unified,chen2015structure}) is to fast locate the most eye-catching objects in a given image. In general, there are two typical methods for the image salient object detection (ISOD) task, including the full convolutional networks (FCNs) based methods and the multi-task learning (MTL) based methods, and we will briefly introduce several most representative methods regarding these two types.

\subsubsection{The FCNs based methods}
The key rationale of the FCNs based methods~\cite{hou2019deeply_dss,wang2019iterative,wu2019cascaded_cpd} is to utilize the multi-scale/multi-level contrast computation to sense saliency clues.
In fact, different network layers usually show different saliency perception abilities, i.e., those deeper layers tend to preserve localization information solely, yet those shallower layers are mainly abundant in tiny details.
Thus, Hou \emph{et al}.~\cite{hou2019deeply_dss} proposed to use short connections between different layers to achieve the multi-scale ISOD, in which the coarse localization information was introduced into the shallower layers, achieving a much improved performance.
Similarly, Wang \emph{et al}.~\cite{wang2019iterative} adopted a top-down and bottom-up inference network, implementing step-by-step optimization via a cooperative and iterative feed-forward and feed-back strategy.
Although these two most representative methods have achieved significant performance improvements, their network structures are generally too heavy. In contrast, Wu \emph{et al}.~\cite{wu2019cascaded_cpd} proposed a lightweight framework, which discarded those high-resolution deep features to speed up detection, of which the motivation is that those deep features in shallower layers usually contribute less to the overall performance yet at high computational costs.

\subsubsection{The MTL based methods}
The key rationale of the MTL based methods is to resort additional auxiliary information to boost the overall performance of the conventional single stream methods, in which such information frequently includes depths~\cite{zhu2019pdnet}, image captions~\cite{zhang2019capsal} and edge clues~\cite{liu2019simple,qin2019basnet,zhao2019egnet}.
Zhu \emph{et al}.~\cite{zhu2019pdnet} proposed to learn a switch map to adaptively fuse the RGB saliency clues with the depth saliency clues to formulate final ISOD result.
Zhang \emph{et al}.~\cite{zhang2019capsal} leveraged the image captions to facilitate their newly proposed weakly supervised ISOD learning scheme, in which the key idea is to utilize the feature similarities between different caption categories to shrink the given problem domain.
Qin \emph{et al}.~\cite{qin2019basnet} proposed a novel edge related loss function to further refine the tiny details in the final ISOD maps.
Similarly, Zhao \emph{et al}.~\cite{zhao2019egnet} combined the edge loss function with multi-level features to further improve the ISOD performance, in which the edge related saliency clues are treated as an explicit indicator to coarsely locate the salient objects.

\subsection{Video Salient Object Detection}
\subsubsection{Conventional hand-crafted methods}
Different to the above mentioned ISOD methods, the video salient object detection (VSOD) is more challenge due to the newly available temporal information.
Previous hand-crafted approaches~\cite{wang2015saliency,guo2017video,chen2018bilevel,guo2019motion,chen2019structure} have widely adopted the low-level saliency clues, which were revealed individually from either spatial branch or temporal branch, to formulate their VSOD.
To fuse spatial and temporal saliency clues, Wang \emph{et al}.~\cite{wang2015saliency} resorted both the spatial edges and the temporal boundaries to facilitate the salient object localization.
Guo \emph{et al}.~\cite{guo2017video} designed a primitive approach to identify the salient object by ranking and selecting the salient proposals.
Chen \emph{et al}.~\cite{chen2018bilevel} devised a bi-level learning strategy to model long-term spatial-temporal saliency consistency. Guo \emph{et al}.~\cite{guo2019motion} proposed a fast VSOD method by using the principal motion vectors to represent the corresponding motion patterns, and such motion message coupling with the color clues together will be fed into the multi-clue optimization framework to achieve the spatiotemporal VSOD.

\subsubsection{Deep-Learning based methods}
The development of convolutional neural networks (CNNs) has fulfilled the needs for performance improvement in the VSOD field.
To date, since the spatial saliency can be measured via the off-the-shelf ISOD deep models, considerable researches have been paid to the measurement of temporal saliency within the deep learning framework, in which the current mainstream works can be categorized into two groups according to their network structures~\cite{cong2018review}, i.e., the single-stream network based methods and the bi-stream network based methods.

We will introduce the single-stream network based methods firstly.
Le \emph{et al}.~\cite{le2017deeply} designed an end-to-end 3D network to directly learn spatiotemporal information.
This 3D framework has added a refinement component at the end of its encoder-decoder backbone network, and its key rationale is to resort the semantical information of the deeper layers to refine its spatiotemporal saliency maps.
Li \emph{et al}.~\cite{li2018flow} developed a novel FCNs based network to conduct VSOD within a stage-wise manner which mainly consists of two main stages; i.e., the spatial saliency maps (using RGB information solely) will be computed in advance, and then those spatial saliency maps within consecutive video frames will be simultaneously fused as the spatiotemporal saliency maps.
To enlarge the temporal sensing scope, Wang et al.~\cite{wang2017video_fcn} adopted the optical flow based correspondences to warp long-term information into the current video frame.
Similarly, Song \emph{et al}.~\cite{song2018pyramid_pdbm} presented a novel scheme to sense the multi-scale spatiotemporal information, in which the key idea is to resort the bi-LSTM network to extract long-term temporal features.
Meanwhile, this work has adopted the pyramid dilated convolutions to extract multi-scale spatial saliency features, which will latterly be fed into the above mentioned bi-LSTM network to achieve the long-term and multi-scale VSOD.
Fan \emph{et al}.~\cite{fan2019shifting_ssav} developed an attention-shift baseline and also released a large-scale saliency-shift-aware dataset for the VSOD problem.

\begin{figure*}[t]
	\centering
	\includegraphics[width=1\linewidth]{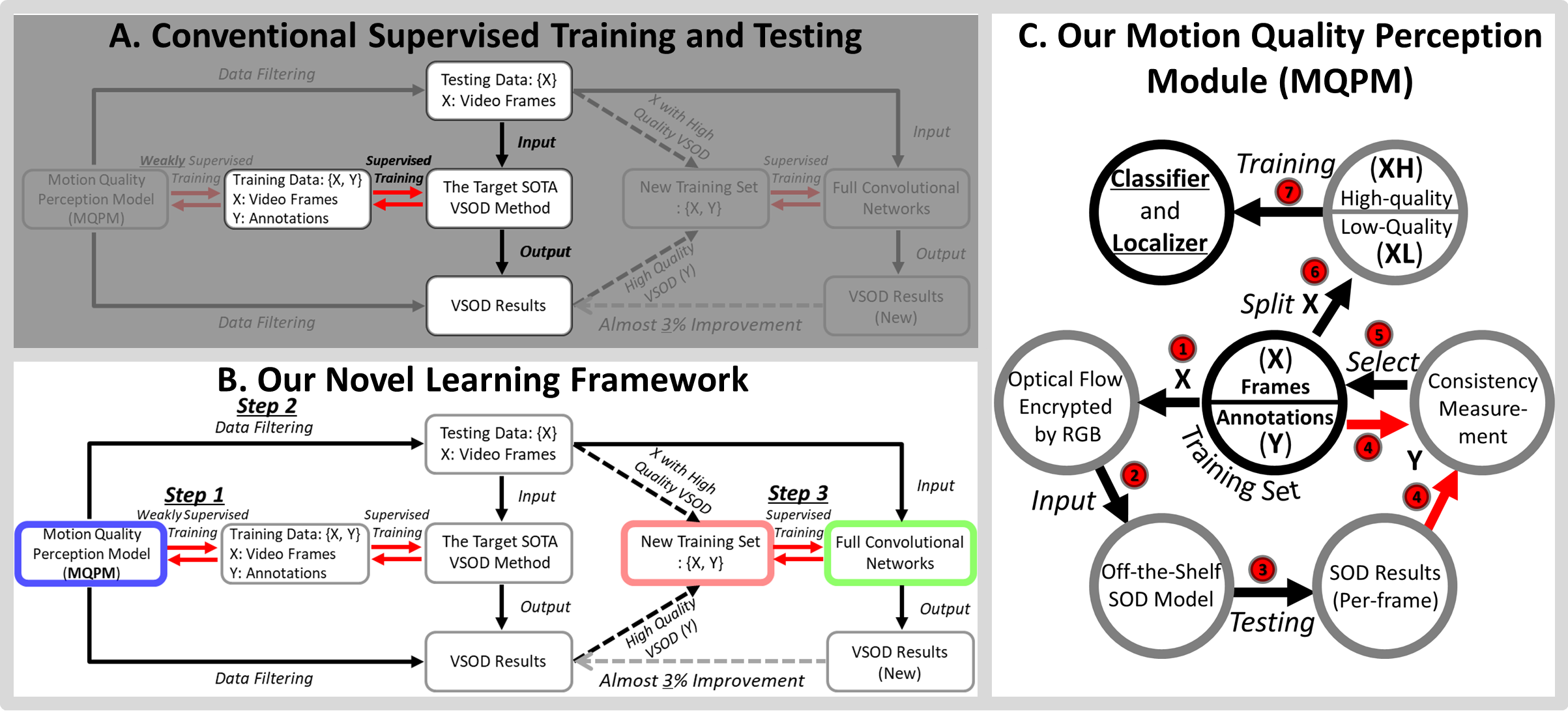}
	\caption{The overall method pipeline. Our novel learning scheme can be applied to the conventional learning scheme (see subfigure-A), and it mainly consists of three steps which have been marked by different colors (red, green and blue) in the subfigure-B; The motion quality perception module is the most important component, and we have demonstrated its details in subfigure-C, where the marks from 1 to 7 respectively show the detailed dataflow.}
	\label{fig:overview}
\end{figure*}

Different from the single-stream networks with limited motion sensing ability~\cite{li2019accurate,ma2019salient}, the bi-stream networks~\cite{chen2020improved,chen2017novel} are usually capable of sensing the motion clues explicitly, in which both the RGB frames and the optical flow maps are treated as the input of their two subbranches, individually.
Then, both the spatial saliency clues and the temporal saliency clues will be computed respectively and latter be fused as the final VSOD results.
Tokmakov \emph{et al}.~\cite{tokmakov2017learning} proposed to feed the concatenated spatial and temporal deep features into the ConvLSTM network, aiming to strike an optimal balance between its temporal branch and spatial branch.
Li \emph{et al}.~\cite{li2019motion_mga} exploited the motion message as attention to boost the overall performance of its spatial branch. Most recently, Gu \emph{et al}.~\cite{gupyramid_pcsa} learned the non-local motion dependencies across several frames, and then it followed the pyramid structure to capture the spatiotemporal saliency clues at various scales.

\section{Proposed Approach}
\subsection{Method Overview}
Given a pre-trained SOTA method (i.e., the target SOTA method), our key idea is to use a subgroup of testing frames with high-quality VSODs to train a novel appearance model, and this novel model will significantly outperform the target SOTA method eventually.
To achieve it, our method mainly consists of three steps, and the detailed method overview can be found in Fig.~\ref{fig:overview}.\\
\underline{\textbf{1)}} Firstly, we weakly train a novel deep model, i.e., the Motion Quality Perception Module (MQPM, blue box).\\
\underline{\textbf{2)}} Next, we use the MQPM to select a subgroup video frames (with high-quality motions) in testing set to formulate a new training set (red box).\\
\underline{\textbf{3)}} Finally, this new training set will be used to train a novel appearance model with much improved VSOD performance (green box).

\subsection{Motion Quality Perception Module}
\label{sec:MQPM}
We demonstrate the detailed MQPM pipeline in Fig.~\ref{fig:overview}-C, the ultimate goal of our approach is to provide a frame-wise binary prediction regarding whether or not the given frame contains high-quality motions.
If yes, it will also provide the spatial locations of the high-quality motions.

To achieve our goal, we should initially divide the training instances (i.e., frames) of the original VSOD training set (i.e., Davis-TR~\cite{perazzi2016benchmark}) into two groups, i.e., one includes frames with high-quality motions, and another one includes frames with low-quality motions only.
Thus, the MQPM can be easily trained by using this partition.

Now the problem is how can we automatically achieve such motion-quality-aware partition in advance.

\begin{figure*}[t]
	\centering
	\includegraphics[width=0.9\linewidth]{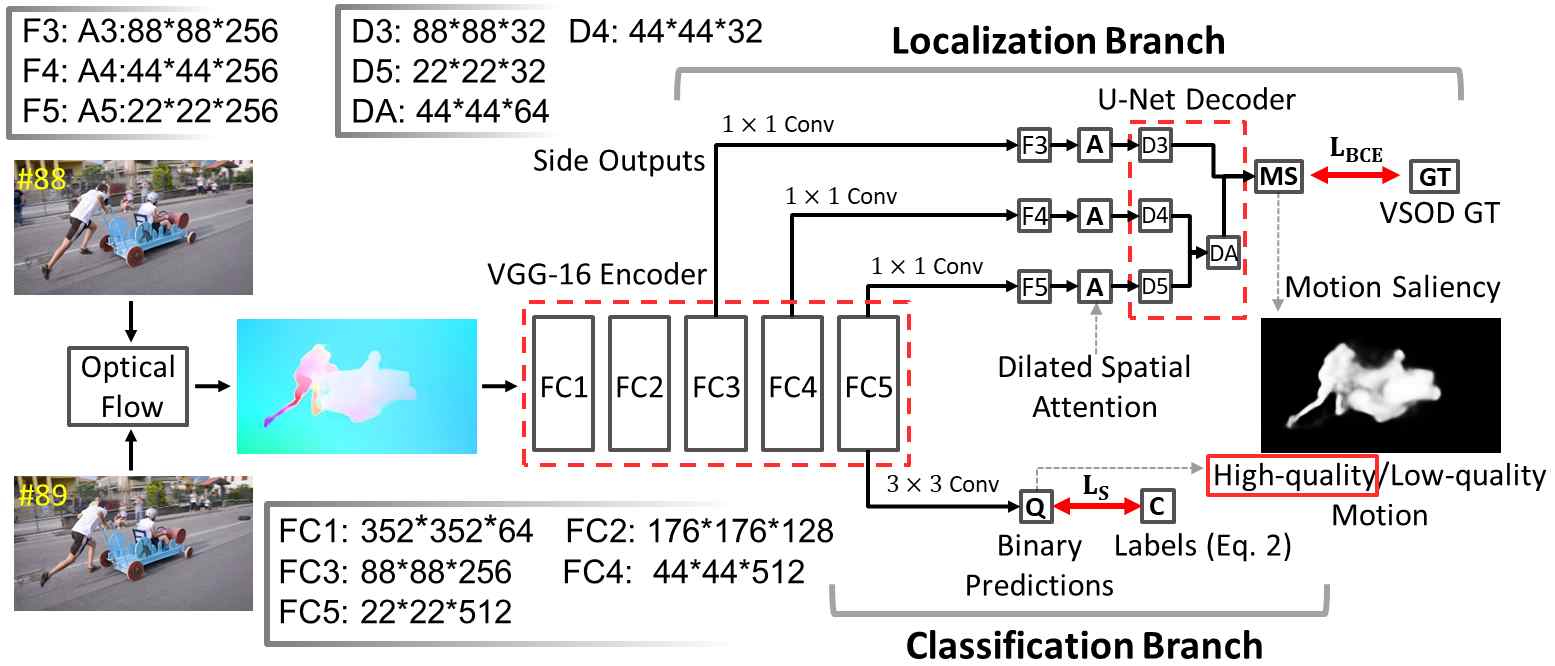}
	\caption{The detailed network architecture of our motion quality perception module (MQPM). For simplicity, we have omitted all up-sampling/down-sampling operations.}
	\label{fig:Net}
\end{figure*}

\vspace{0.2cm}
\subsubsection{Motion Quality Measurement}
As is shown in the 2nd row of Fig.~\ref{fig:motivation}, we have demonstrated the corresponding optical flow results (encrypted using RGB color) of some consecutive frames in a given video sequence(i.e., the ``\emph{worm}'' sequence from the widely-used Davis set).
Notice that these optical flow results are computed by using the off-the-shelf optical flow tool~\cite{sun2018pwc} to sense motions between two consecutive frames, in which the RGB colors at different pixels denote the estimated motion intensities and directions.
It can be easily observed in Fig.~\ref{fig:motivation} that the video frames with high-quality motions (e.g., the frame \#18) usually share some distinct attributes in common, i.e., the optical flow values inside the salient object (i.e., the worm) will be totally different to the non-salient surroundings nearby.
Based on this, we propose a simple yet effective way to measure the motion quality score (MQS) as Eq.~\ref{eq:MQS} with a quite straight froward rationale; i.e., the salient objects in those frames with high-quality motions will have large probability to be successfully detected by the off-the-shelf image salient object detection method, and these frames should be assigned with large MQSs.
\begin{equation}
\large
\rm MQS_\emph{i} = \emph{\Large f}\Big(\Theta(OF_\emph{i}),GT_\emph{i}\Big),
\label{eq:MQS}
\end{equation}
where $\rm OF_\emph{i}$ denotes the optical flow result of the $i$-th frame, and $\rm GT$ denotes the human well-annotated pixel-wise VSOD saliency ground truth; $\Theta$ denotes a pre-trained image salient object detection deep model, which we choose the off-the-shelf CPD~\cite{wu2019cascaded_cpd} due to its lightweight implementation; $f$ denotes the consistency measurement between the SOD made by $\Theta$ and the $\rm GT$.
In fact, there are various consistency measurements which are widely used to conduct quantitative evaluations, such as MAE~\cite{perazzi2012saliency}, F-Measure~\cite{achanta2009frequency} and S-Measure~\cite{fan2017structure}.
For simplicity, we choose the S-Measure as the consistency measurement $f$ in Eq.~\ref{eq:MQS}.
Notice that we have also tested other measurements, but the overall performance won't change much, i.e., floating of two decimal places mostly.

\vspace{0.2cm}
\subsubsection{Training Set for MQPM}
To train our MQPM, we need to weakly assign binary labels for each frame in the Davis training set regarding whether it contains high-quality motions.
Therefore, we use the motion quality scores (MQS, Eq.~\ref{eq:MQS}) as the key indicator to produce such labels ($\rm Label_\emph{i}$) as following:
\begin{equation}
\large
\rm
Label_\emph{i} = \left\{ \begin{array}{ll}
\rm 0\ \ \ \ \emph{if}\ \ MQS_\emph{i}<\lambda\ \\
\ \ \ \ \ \ \ \ \ \ (i.e., {\rm XL}\ in\ Fig.~\ref{fig:overview}C)\\
\rm 1\ \ \ \ \emph{otherwise} \\
\ \ \ \ \ \ \ \ \ \ (i.e., {\rm XH}\ in\ Fig.~\ref{fig:overview}C)\\
\end{array}\right.,
\label{eq:PatchScore}
\end{equation}
where XH means high-quality optical flow frames, and XL denotes low-quality movements. When motion quality scores (MQS) is less than the threshold value $\lambda$, the label is assigned to 1. Otherwise, the label is assigned to 0, where $\lambda$ is a pre-defined decision threshold.
To ensure an optimal balance between positive-1 and negative-0 training instances, we iteratively update $\lambda$ until the convergence via Eq.~\ref{eq:OM} and Eq.~\ref{eq:th}.
\begin{equation}
\large
\rm \omega = \frac{\int_\lambda^\infty MQS\cdot \emph{P}(MQS)\ \ \emph{d}(MQS)}{\int_\lambda^\infty \emph{P}(MQS)\ \ \emph{d}(MQS)},
\label{eq:OM}
\end{equation}
\begin{equation}
\large
\lambda = (1+\omega)/2,
\label{eq:th}
\end{equation}
where $P({\rm MQS})$ is the probability distribution of MQS in the entire VSOD training set.

Thus far, we can formulate the training set as $\{\rm X_\emph{i}$, $GT_\emph{i}$, $Label_\emph{i}\}$, where $\rm X_\emph{i}$ denotes the $i$-th video frame, $\rm GT$ is the original binary VSOD ground truth.
Next, we will introduce how to train the MQPM by using this training set.

\vspace{0.2cm}
\subsubsection{MQPM Training}
We formulate our MQPM training as a multi-task procedure following the vanilla bi-stream structure, in which one stream aims the binary motion quality prediction (i.e., classification) and another stream conducts the pixel-wise motion saliency detection (i.e., localization).

As is shown in Fig.~\ref{fig:Net}, the MQPM takes the RGB encrypted optical flow data as input, and its output comprises two parts: \underline{\textbf{1)}} motion saliency map; \underline{\textbf{2)}} motion quality prediction.
The main network structure of MQPM comprises three components: one feature encoder (VGG-16~\cite{simonyan2014very}) and two sub-branches with different loss functions.

The motion saliency branch takes the last three encoder layers as input.
Next, each of these input will be fed into the widely-used multi-scale dilated attention module (with dilation factors ranging between \{2,4,6,8\}) to filter those irrelevant features.
Thus, the motion saliency map can be computed by applying the U-Net~\cite{ronneberger2015u} decoder iteratively, in which the binary cross entropy loss ($\rm L_{BCE}$) is used.
Meanwhile, the classification branch only takes the last decoder layer as input.
Thus, the total loss function can be represented as Eq.~\ref{eq:jointloss}.
\begin{equation}
\large
\rm L_{total} = L_{BCE}+L_{S},
\label{eq:jointloss}
\end{equation}
where the binary cross entropy loss ($\rm L_{BCE}$) can be detailed as Eq.~\ref{eq:BCE}, and the $\rm L_{S}$ is a typical binary classification loss as is shown in Eq.~\ref{eq:SoftMax}.
\begin{equation}
\large
\begin{split}
\rm &L_{BCE}=\rm -\sum_{\emph{i}}\sum_{\emph{u}} GT_\emph{i}(\emph{u})\times \log\Big(MS_\emph{i}(\emph{u})\Big)\\
&\rm -\sum_{\emph{i}}\sum_{\emph{u}} \Big(1-GT_\emph{i}(\emph{u})\Big)\times \log\Big(1-MS_\emph{i}(\emph{u})\Big),
\end{split}
\label{eq:BCE}
\end{equation}
where $\rm MS_\emph{i}(\emph{u})$ denotes the predicted motion saliency value at the $u$-th pixel in the $i$-th frame; $GT_\emph{i}(\emph{u})$ represents ground truth value at the $u$-th pixel in the $i$-th frame; ``$\times$'' is a conventional multiplication operation; $\log$ is a typical logarithmic mathematical operation.
\begin{equation}
\large
\begin{split}
\rm L_{S}=-\sum_\emph{i}& \Big[Label_\emph{i}\times\log Q_\emph{i}\\
&+(1-Label_\emph{i})\times\log(1- Q_\emph{i})\Big],
\end{split}
\label{eq:SoftMax}
\end{equation}
where $\rm L_{S}$ is a logistic regression cost loss function; $\emph{Q}_\emph{i}$ denote the confidence regarding the category predictions (i.e., high-quality/low-quality motions); $\rm Label_\emph{i}$ is the previously determined motion quality label (Eq.~\ref{eq:PatchScore}).

\subsection{New Training Set For VSOD}
\subsubsection{Initialization}
Thus far, the motion quality perception module (MQPM) has been trained, providing two vital information which can be used to improve the target SOTA method:
\underline{\textbf{1)}} the binary motion quality prediction; \underline{\textbf{2)}} the motion saliency map.

\begin{figure*}[t]
	\centering
	\includegraphics[width=\linewidth]{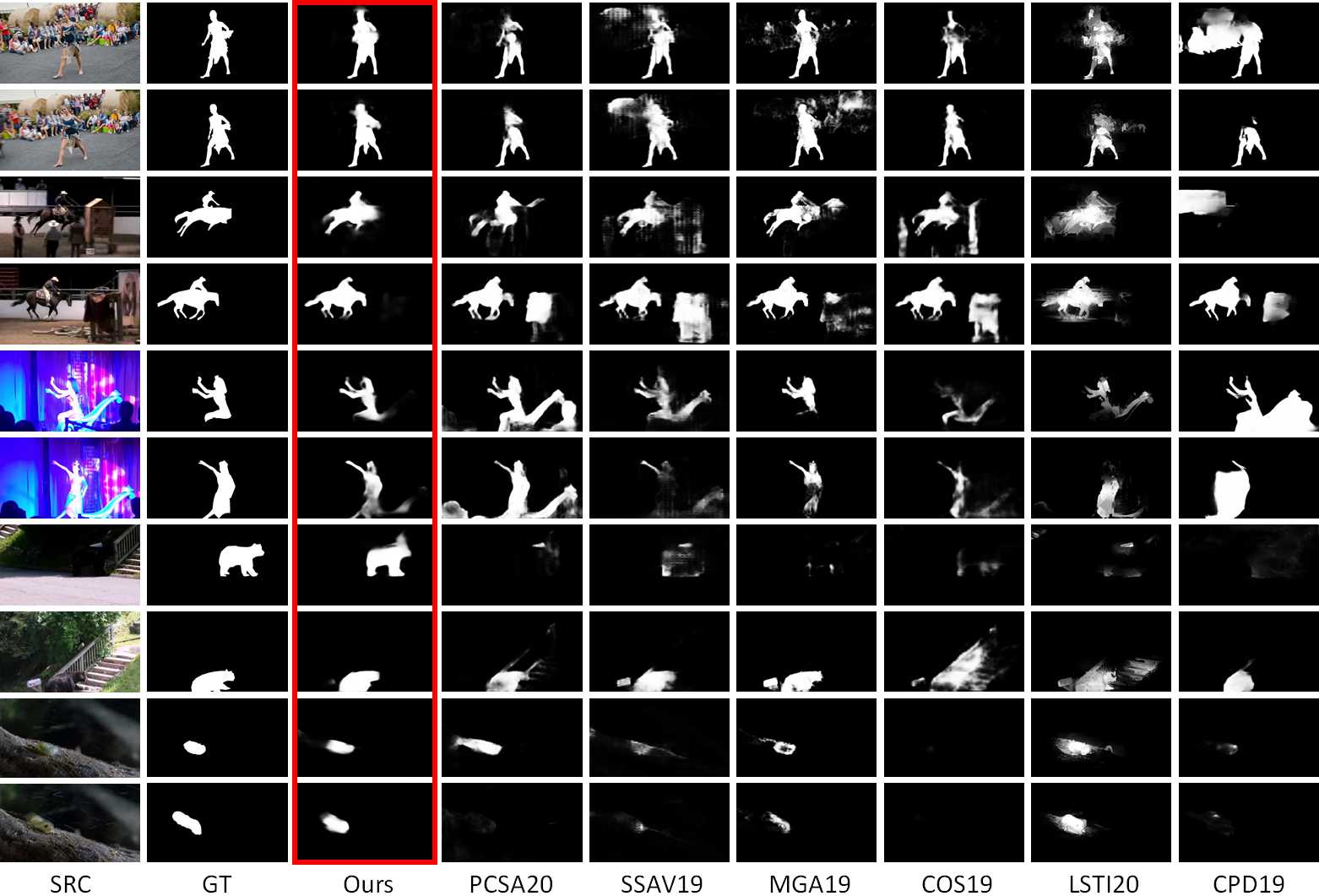}
	\caption{Qualitative comparisons with the current SOTA methods. Due to the limited space, we only list six most representative ones here, including PCSA20~\cite{gupyramid_pcsa}, SSAV19~\cite{fan2019shifting_ssav}, MGA19~\cite{li2019motion_mga}, COS19~\cite{lu2019see}, LSTI20~\cite{chen2019improved} and CPD19~\cite{wu2019cascaded_cpd}.}
	\label{fig:result end vision}
\end{figure*}

As we have mentioned before, the former one can be used as an explicit indicator to tell which frames in the VSOD testing set should be selected, while the latter will be used as a double-check to ensure the selected frames are really with high-quality motions which are capable of benefiting the VSOD training in practice.
Here we will use both of these two to facilitate the construction of a new training set, which only comprises video frames containing high-quality motions.
And this new training set will be used to start a new round of network training and improve the target SOTA method performance eventually.

For each frame in the VSOD testing set, we first compute its optical flow results frame-by-frame, and then feed these optical flow results into the well-trained MQPM, and thus those frames (i.e., the original frames rather than their optical flow results) which are predicted to have contained ``High-quality Motions'' will be directly pooled as the initial version of the new training set.

Next, for each training instance ($\{\textbf{X}_i,\textbf{Y}_i\}$) in this new training set, it mainly consists of two components, including the original frame \textbf{X} and the corresponding VSOD result predicted by the target SOTA method (trained using both spatial and temporal information) as its training objective (i.e., \textbf{Y}, see the pictorial demonstration in the red box of Fig.~\ref{fig:overview}).

Also, it is worthy mentioning that we can not directly use the motion saliency maps (i.e., the output of the localization branch in Fig.~\ref{fig:Net}) as the training objectives.
The main reason is that the motion saliency maps are usually with blur object boundaries (due to absent of spatial information), and thus the performance improvement may be severally limited if we directly apply the motion saliency maps as the pseudo-GT during this new round training, and the corresponding quantitative evidences can be found in Table.~\ref{tab:SC}.

\begin{table*}[t]
	\centering
	\caption{Ablation study regrading our data filtering strategy, where baseline denotes the target SOTA method (i.e., SSAV), see more details in Sec.~\ref{sec:AS}.}
	\resizebox{1\textwidth}{!}{
		\begin{tabular}{|l|c|c|c|c|c|c|c|c|c|c|c|c|c|c|c|}
			\toprule
			Dataset & \multicolumn{3}{c|}{Davis} & \multicolumn{3}{c|}{Segv2} & \multicolumn{3}{c|}{Visal} & \multicolumn{3}{c|}{DAVSOD} & \multicolumn{3}{c|}{VOS} \\
			\midrule
			Metric & \multicolumn{1}{c}{maxF} & \multicolumn{1}{c}{S-M} & MAE   & \multicolumn{1}{c}{maxF} & \multicolumn{1}{c}{S-M} & MAE   & \multicolumn{1}{c}{maxF} & \multicolumn{1}{c}{S-M} & MAE   & \multicolumn{1}{c}{maxF} & \multicolumn{1}{c}{S-M} & MAE   & \multicolumn{1}{c}{maxF} & \multicolumn{1}{c}{S-M} & MAE \\
			\midrule
			Basline & \multicolumn{1}{c}{0.861} & \multicolumn{1}{c}{0.893} & 0.028  & \multicolumn{1}{c}{0.801} & \multicolumn{1}{c}{0.851} & 0.023  & \multicolumn{1}{c}{0.939} & \multicolumn{1}{c}{0.943} & 0.020  & \multicolumn{1}{c}{0.603} & \multicolumn{1}{c}{0.724} & 0.092  & \multicolumn{1}{c}{0.742} & \multicolumn{1}{c}{0.819} & 0.073  \\
			T=1   & \multicolumn{1}{c}{0.892} & \multicolumn{1}{c}{\textbf{0.910}} & 0.019  & \multicolumn{1}{c}{0.826} & \multicolumn{1}{c}{0.873} & 0.020  & \multicolumn{1}{c}{0.934} & \multicolumn{1}{c}{0.939} & 0.018  & \multicolumn{1}{c}{0.686} & \multicolumn{1}{c}{0.764} & 0.076  & \multicolumn{1}{c}{0.755} & \multicolumn{1}{c}{0.820} & 0.067  \\
			T=1/2 & \multicolumn{1}{c}{0.890} & \multicolumn{1}{c}{0.908} & 0.018  & \multicolumn{1}{c}{0.824} & \multicolumn{1}{c}{0.866} & 0.019  & \multicolumn{1}{c}{0.934} & \multicolumn{1}{c}{0.935} & 0.018  & \multicolumn{1}{c}{0.686} & \multicolumn{1}{c}{0.760} & 0.074  & \multicolumn{1}{c}{0.760} & \multicolumn{1}{c}{0.819} & 0.066  \\
			T=1/3 & \multicolumn{1}{c}{0.889} & \multicolumn{1}{c}{0.906} & 0.020  & \multicolumn{1}{c}{0.833} & \multicolumn{1}{c}{0.874} & 0.019  & \multicolumn{1}{c}{0.938} & \multicolumn{1}{c}{\textbf{0.942}} & 0.016  & \multicolumn{1}{c}{0.696} & \multicolumn{1}{c}{0.769} & 0.072  & \multicolumn{1}{c}{0.758} & \multicolumn{1}{c}{0.825} & 0.064  \\
			T=1/4 & \multicolumn{1}{c}{0.893} & \multicolumn{1}{c}{0.908} & \textbf{0.018} & \multicolumn{1}{c}{0.832} & \multicolumn{1}{c}{0.870} & \textbf{0.018} & \multicolumn{1}{c}{0.934} & \multicolumn{1}{c}{0.933} & \textbf{0.016} & \multicolumn{1}{c}{0.693} & \multicolumn{1}{c}{0.768} & 0.074  & \multicolumn{1}{c}{0.756} & \multicolumn{1}{c}{0.822} & \textbf{0.063} \\
			T=1/5 & \multicolumn{1}{c}{\textbf{0.894}} & \multicolumn{1}{c}{0.906} & 0.020  & \multicolumn{1}{c}{\textbf{0.836}} & \multicolumn{1}{c}{\textbf{0.880}} & 0.019  & \multicolumn{1}{c}{0.935} & \multicolumn{1}{c}{0.940} & 0.017  & \multicolumn{1}{c}{\textbf{0.699}} & \multicolumn{1}{c}{\textbf{0.774}} & \textbf{0.071} & \multicolumn{1}{c}{\textbf{0.767}} & \multicolumn{1}{c}{\textbf{0.831}} & 0.066  \\
			T=1/10 & \multicolumn{1}{c}{0.888} & \multicolumn{1}{c}{0.906} & 0.019  & \multicolumn{1}{c}{0.836} & \multicolumn{1}{c}{0.876} & 0.018  & \multicolumn{1}{c}{\textbf{0.940}} & \multicolumn{1}{c}{0.939} & 0.016  & \multicolumn{1}{c}{0.698} & \multicolumn{1}{c}{0.769} & 0.071  & \multicolumn{1}{c}{0.738} & \multicolumn{1}{c}{0.812} & 0.070  \\
			\bottomrule
		\end{tabular}%
	}
	\label{tab:T}%
\end{table*}%

\begin{table*}[t]
	\centering
	\caption{Proofs regarding the effectiveness of our motion quality perception module (MQPM). The quantitative metrics include the maxF (larger is better), meanF (larger is better), adpF (larger is better), S-Measure (larger is better) and MAE (smaller is better). By using the MQPM as the indicator, those frames which are predicted to have high-quality motions can outperform other frames significantly, in which we choose the SSAV~\cite{fan2019shifting_ssav} as the target SOTA method for example here.}
	\vspace{-0.1cm}
	\resizebox{0.8\textwidth}{!}{
		\begin{tabular}{|r|ccccc|c|c|c|c|c|}
			\toprule
			Quality & \multicolumn{5}{c|}{Frames with High-quality Motions (HQ)}               & \multicolumn{5}{c|}{Frames with Low-quality Motions (LQ)} \\
			\midrule
			Metric & maxF  & meanF & adpF  & S-M   & MAE   & \multicolumn{1}{c}{maxF} & \multicolumn{1}{c}{meanF} & \multicolumn{1}{c}{adpF} & \multicolumn{1}{c}{S-M} & MAE \\
			\midrule
			Davis~\cite{perazzi2016benchmark} & 0.884  & 0.840  & 0.800  & 0.906  & 0.022  & \multicolumn{1}{c}{0.828} & \multicolumn{1}{c}{0.782} & \multicolumn{1}{c}{0.719} & \multicolumn{1}{c}{0.875} & 0.034  \\
			Segv2~\cite{li2013video} & 0.864  & 0.808  & 0.834  & 0.881  & 0.024  & \multicolumn{1}{c}{0.780} & \multicolumn{1}{c}{0.726} & \multicolumn{1}{c}{0.709} & \multicolumn{1}{c}{0.852} & 0.024  \\
			DAVSOD~\cite{fan2019shifting_ssav} & 0.653  & 0.621  & 0.626  & 0.753  & 0.080  & \multicolumn{1}{c}{0.642} & \multicolumn{1}{c}{0.611} & \multicolumn{1}{c}{0.611} & \multicolumn{1}{c}{0.738} & 0.086  \\
			Visal~\cite{wang2015consistent} & 0.883  & 0.850  & 0.832  & 0.910  & 0.025  & \multicolumn{1}{c}{0.938} & \multicolumn{1}{c}{0.895} & \multicolumn{1}{c}{0.841} & \multicolumn{1}{c}{0.945} & 0.014 \\
			VOS~\cite{li2017benchmark}   & 0.767  & 0.739  & 0.749  & 0.815  & 0.073  & \multicolumn{1}{c}{0.734} & \multicolumn{1}{c}{0.697} & \multicolumn{1}{c}{0.700} & \multicolumn{1}{c}{0.816} & 0.074 \\
			\midrule
			Total & 0.810  & 0.772  & 0.768  & 0.853  & 0.045  & \multicolumn{1}{c}{0.784} & \multicolumn{1}{c}{0.742} & \multicolumn{1}{c}{0.716} & \multicolumn{1}{c}{0.845} & 0.046 \\
			\bottomrule
		\end{tabular}%
	}
	\label{tab:HL}%
\end{table*}%

\subsubsection{Data Filtering}
\label{sec:DF}
As we have mentioned before, our rationale is based on the assumption that the SOTA methods tend to exhibit high-quality VSOD over those frames with high-quality motions (see the quantitative proofs in Table.~\ref{tab:HL}).
In fact, this assumption holds in most cases.
However, there still exists exceptions occasionally.

As is shown in Fig.~\ref{fig:motivation}, our MQPM has predicted that the \#20 frame has large probability of containing some high-quality motions, and the optical flow result of the \#20 frame (in the 2nd row) is indeed capable of separating the salient object from its non-salient surroundings nearby, producing high-quality motion saliency map as well (in the 3rd row).
However, the VSOD predicted by the target SOTA method (i.e., it can be any SOTA method, here, we simply choose the SSAV~\cite{fan2019shifting_ssav} for example) failed to completely detect the salient object, and it may degrade the overall performance if the new training set contains a large number of such cases.

Meanwhile, we have noticed that there exists a large number of consecutive frames in the testing VSOD set (almost 30\%) which are tend to be predicted as the ones containing high-quality motions.
Since these consecutive frames usually share similar spatial appearance in general, it will easily lead to an over-fitted appearance model if we use all these frames during the up-coming training.

\begin{table*}[t]
	\centering
	\caption{Component quantitative evaluation results. The quantitative metrics include the maxF (larger is better), S-Measure (larger is better) and MAE (smaller is better), more details can be found in Sec.~\ref{sec:CE}.}
	\vspace{-0.1cm}
	\resizebox{1\textwidth}{!}{
		\begin{tabular}{|c|c|c|c|c|c|c|c|c|c|c|c|c|c|c|c|}
			\toprule
			Dataset & \multicolumn{3}{c|}{Davis} & \multicolumn{3}{c|}{Segv2} & \multicolumn{3}{c|}{Visal} & \multicolumn{3}{c|}{DAVSOD} & \multicolumn{3}{c|}{VOS} \\
			\midrule
			Metric & \multicolumn{1}{c}{maxF} & \multicolumn{1}{c}{S-M} & MAE   & \multicolumn{1}{c}{maxF} & \multicolumn{1}{c}{S-M} & MAE   & \multicolumn{1}{c}{maxF} & \multicolumn{1}{c}{S-M} & MAE   & \multicolumn{1}{c}{maxF} & \multicolumn{1}{c}{S-M} & MAE   & \multicolumn{1}{c}{maxF} & \multicolumn{1}{c}{S-M} & MAE \\
			\midrule
			MS Baseline   & \multicolumn{1}{c}{0.798} & \multicolumn{1}{c}{0.854} & 0.044 & \multicolumn{1}{c}{0.648} & \multicolumn{1}{c}{0.760} & 0.054 & \multicolumn{1}{c}{0.627} & \multicolumn{1}{c}{0.738} & 0.079 & \multicolumn{1}{c}{0.450} & \multicolumn{1}{c}{0.613} & 0.148 & \multicolumn{1}{c}{0.405} & \multicolumn{1}{c}{0.566} & 0.167 \\
			MS$-$MQPM  & \multicolumn{1}{c}{0.784} & \multicolumn{1}{c}{0.844} & 0.043 & \multicolumn{1}{c}{0.656} & \multicolumn{1}{c}{0.761} & 0.053 & \multicolumn{1}{c}{0.688} & \multicolumn{1}{c}{0.774} & 0.075 & \multicolumn{1}{c}{0.488} & \multicolumn{1}{c}{0.632} & 0.143 & \multicolumn{1}{c}{0.501} & \multicolumn{1}{c}{0.617} & 0.161 \\
			MS$+$MQPM & \multicolumn{1}{c}{0.814} & \multicolumn{1}{c}{0.866} & 0.032 & \multicolumn{1}{c}{0.760} & \multicolumn{1}{c}{0.832} & 0.028 & \multicolumn{1}{c}{0.745} & \multicolumn{1}{c}{0.809} & 0.051 & \multicolumn{1}{c}{0.569} & \multicolumn{1}{c}{0.685} & 0.107 & \multicolumn{1}{c}{0.627} & \multicolumn{1}{c}{0.702} & 0.108 \\
			MS$+$MQPM$+$SOTA & \multicolumn{1}{c}{\textbf{0.894}} & \multicolumn{1}{c}{\textbf{0.906}} & \textbf{0.020} & \multicolumn{1}{c}{\textbf{0.836}} & \multicolumn{1}{c}{\textbf{0.880}} & \textbf{0.019} & \multicolumn{1}{c}{\textbf{0.935}} & \multicolumn{1}{c}{\textbf{0.940}} & \textbf{0.017} & \multicolumn{1}{c}{\textbf{0.699}} & \multicolumn{1}{c}{\textbf{0.774}} & \textbf{0.071} & \multicolumn{1}{c}{\textbf{0.767}} & \multicolumn{1}{c}{\textbf{0.831}} & \textbf{0.066} \\
			\bottomrule
		\end{tabular}%
	}
	\label{tab:SC}%
\end{table*}%

%
\begin{table*}[!t]
	\centering
	\caption{Quantitative comparisons with current SOTA methods. The top three results are marked by red, green and blue, respectively.}
	\resizebox{0.9\textwidth}{!}{
		\begin{tabular}{|r|c|c|c|c|c|c|c|c|c|c|c|c|c|c|}
			\toprule
			\multirow{3}[4]{*}{Dataset} & \multirow{3}[4]{*}{Metric} & \multirow{3}[3]{*}{Ours} & \multicolumn{2}{c|}{2020} & \multicolumn{4}{c|}{2019}     & \multicolumn{3}{c|}{2018} & \multicolumn{3}{c|}{2017} \\
			\cmidrule{4-15}          &       &       & PCSA  & LSTI  & SSAV  & MGA   & COS   & CPD   & PDBM  & MBNM  & SCOM  & SFLR  & SGSP  & STBP \\
			&       &       & \cite{gupyramid_pcsa}    & \cite{chen2019improved}    & \cite{fan2019shifting_ssav}    & \cite{li2019motion_mga}    & \cite{lu2019see}    & \cite{wu2019cascaded_cpd}    & \cite{song2018pyramid_pdbm}    & \cite{li2018unsupervised}    & \cite{xi2016salient}    & \cite{chen2017video}    & \cite{liu2016saliency}    & \cite{xi2016salient} \\
			\midrule
			\multirow{3}[2]{*}{Davis~\cite{perazzi2016benchmark}} & maxF  & \textcolor[rgb]{1.000, 0.000, 0.000}{\textbf{0.894}} & \multicolumn{1}{c}{\textcolor[rgb]{0.000, 0.439, 0.753}{\textbf{0.880}}} & 0.850  & \multicolumn{1}{c}{0.861} & \multicolumn{1}{c}{\textcolor[rgb]{0.000, 0.690, 0.314}{\textbf{0.892}}} & \multicolumn{1}{c}{0.875} & 0.778  & \multicolumn{1}{c}{0.855} & \multicolumn{1}{c}{0.861} & 0.783  & \multicolumn{1}{c}{0.727} & \multicolumn{1}{c}{0.655} & 0.544  \\
			& S-M   & \textcolor[rgb]{0.000, 0.690, 0.314}{\textbf{0.906}} & \multicolumn{1}{c}{\textcolor[rgb]{0.000, 0.439, 0.753}{\textbf{0.902}}} & 0.876  & \multicolumn{1}{c}{0.893} & \multicolumn{1}{c}{\textcolor[rgb]{1.000, 0.000, 0.000}{\textbf{0.910}}} & \multicolumn{1}{c}{\textcolor[rgb]{0.000, 0.439, 0.753}{\textbf{0.902}}} & 0.859  & \multicolumn{1}{c}{0.882} & \multicolumn{1}{c}{0.887} & 0.832  & \multicolumn{1}{c}{0.790} & \multicolumn{1}{c}{0.692} & 0.677  \\
			& MAE   & \textcolor[rgb]{1.000, 0.000, 0.000}{\textbf{0.020}} & \multicolumn{1}{c}{\textcolor[rgb]{0.000, 0.690, 0.314}{\textbf{0.022}}} & 0.034  & \multicolumn{1}{c}{\textcolor[rgb]{0.000, 0.439, 0.753}{\textbf{0.023}}} & \multicolumn{1}{c}{\textcolor[rgb]{0.000, 0.439, 0.753}{\textbf{0.023}}} & \multicolumn{1}{c}{\textcolor[rgb]{1.000, 0.000, 0.000}{\textbf{0.020}}} & 0.032  & \multicolumn{1}{c}{0.028} & \multicolumn{1}{c}{0.031} & 0.064  & \multicolumn{1}{c}{0.056} & \multicolumn{1}{c}{0.138} & 0.096  \\
			\midrule
			\multirow{3}[2]{*}{SegV2~\cite{li2013video}} & maxF  & \textcolor[rgb]{0.000, 0.690, 0.314}{\textbf{0.836}} & \multicolumn{1}{c}{0.810} & \textcolor[rgb]{1.000, 0.000, 0.000}{\textbf{0.858}} & \multicolumn{1}{c}{0.801} & \multicolumn{1}{c}{\textcolor[rgb]{0.000, 0.439, 0.753}{\textbf{0.821}}} & \multicolumn{1}{c}{0.801} & 0.778  & \multicolumn{1}{c}{0.800} & \multicolumn{1}{c}{0.716} & 0.764  & \multicolumn{1}{c}{0.745} & \multicolumn{1}{c}{0.673} & 0.640  \\
			& S-M   & \textcolor[rgb]{1.000, 0.000, 0.000}{\textbf{0.880}} & \multicolumn{1}{c}{\textcolor[rgb]{0.000, 0.439, 0.753}{\textbf{0.865}}} & \textcolor[rgb]{0.000, 0.690, 0.314}{\textbf{0.870}} & \multicolumn{1}{c}{0.851} & \multicolumn{1}{c}{\textcolor[rgb]{0.000, 0.439, 0.753}{\textbf{0.865}}} & \multicolumn{1}{c}{0.850} & 0.841  & \multicolumn{1}{c}{0.864} & \multicolumn{1}{c}{0.809} & 0.815  & \multicolumn{1}{c}{0.804} & \multicolumn{1}{c}{0.681} & 0.735  \\
			& MAE   & \textcolor[rgb]{1.000, 0.000, 0.000}{\textbf{0.019}} & \multicolumn{1}{c}{0.025} & 0.025  & \multicolumn{1}{c}{\textcolor[rgb]{0.000, 0.439, 0.753}{\textbf{0.023}}} & \multicolumn{1}{c}{0.030} & \multicolumn{1}{c}{\textcolor[rgb]{0.000, 0.690, 0.314}{\textbf{0.020}}} & \textcolor[rgb]{0.000, 0.439, 0.753}{\textbf{0.023}} & \multicolumn{1}{c}{0.024} & \multicolumn{1}{c}{0.026} & 0.030  & \multicolumn{1}{c}{0.037} & \multicolumn{1}{c}{0.124} & 0.061  \\
			\midrule
			\multirow{3}[2]{*}{Visal~\cite{wang2015consistent}} & maxF  & 0.935  & \multicolumn{1}{c}{\textcolor[rgb]{0.000, 0.439, 0.753}{\textbf{0.940}}} & 0.905  & \multicolumn{1}{c}{0.939} & \multicolumn{1}{c}{0.933} & \multicolumn{1}{c}{\textcolor[rgb]{1.000, 0.000, 0.000}{\textbf{0.966}}} & \textcolor[rgb]{0.000, 0.690, 0.314}{\textbf{0.941}} & \multicolumn{1}{c}{0.888} & \multicolumn{1}{c}{0.883} & 0.831  & \multicolumn{1}{c}{0.779} & \multicolumn{1}{c}{0.677} & 0.622  \\
			& S-M   & 0.940  & \multicolumn{1}{c}{\textcolor[rgb]{0.000, 0.690, 0.314}{\textbf{0.946}}} & 0.916  & \multicolumn{1}{c}{\textcolor[rgb]{0.000, 0.439, 0.753}{\textbf{0.943}}} & \multicolumn{1}{c}{0.936} & \multicolumn{1}{c}{\textcolor[rgb]{1.000, 0.000, 0.000}{\textbf{0.965}}} & 0.942  & \multicolumn{1}{c}{0.907} & \multicolumn{1}{c}{0.898} & 0.762  & \multicolumn{1}{c}{0.814} & \multicolumn{1}{c}{0.706} & 0.629  \\
			& MAE   & \textcolor[rgb]{0.000, 0.439, 0.753}{\textbf{0.017}} & \multicolumn{1}{c}{\textcolor[rgb]{0.000, 0.439, 0.753}{\textbf{0.017}}} & 0.033  & \multicolumn{1}{c}{0.020} & \multicolumn{1}{c}{\textcolor[rgb]{0.000, 0.439, 0.753}{\textbf{0.017}}} & \multicolumn{1}{c}{\textcolor[rgb]{1.000, 0.000, 0.000}{\textbf{0.011}}} & \textcolor[rgb]{0.000, 0.690, 0.314}{\textbf{0.016}} & \multicolumn{1}{c}{0.032} & \multicolumn{1}{c}{0.020} & 0.122  & \multicolumn{1}{c}{0.062} & \multicolumn{1}{c}{0.165} & 0.163  \\
			\midrule
			\multirow{3}[2]{*}{DAVSOD~\cite{fan2019shifting_ssav}} & maxF  & \textcolor[rgb]{1.000, 0.000, 0.000}{\textbf{0.699}} & \multicolumn{1}{c}{\textcolor[rgb]{0.000, 0.690, 0.314}{\textbf{0.655}}} & 0.585  & \multicolumn{1}{c}{0.603} & \multicolumn{1}{c}{\textcolor[rgb]{0.000, 0.439, 0.753}{\textbf{0.640}}} & \multicolumn{1}{c}{0.614} & 0.608  & \multicolumn{1}{c}{0.572} & \multicolumn{1}{c}{0.520} & 0.464  & \multicolumn{1}{c}{0.478} & \multicolumn{1}{c}{0.426} & 0.410  \\
			& S-M   & \textcolor[rgb]{1.000, 0.000, 0.000}{\textbf{0.774}} & \multicolumn{1}{c}{\textcolor[rgb]{0.000, 0.690, 0.314}{\textbf{0.741}}} & 0.695  & \multicolumn{1}{c}{0.724} & \multicolumn{1}{c}{\textcolor[rgb]{0.000, 0.439, 0.753}{\textbf{0.738}}} & \multicolumn{1}{c}{0.725} & 0.724  & \multicolumn{1}{c}{0.698} & \multicolumn{1}{c}{0.637} & 0.599  & \multicolumn{1}{c}{0.624} & \multicolumn{1}{c}{0.577} & 0.568  \\
			& MAE   & \textcolor[rgb]{1.000, 0.000, 0.000}{\textbf{0.071}} & \multicolumn{1}{c}{\textcolor[rgb]{0.000, 0.439, 0.753}{\textbf{0.086}}} & 0.106  & \multicolumn{1}{c}{0.092} & \multicolumn{1}{c}{\textcolor[rgb]{0.000, 0.690, 0.314}{\textbf{0.084}}} & \multicolumn{1}{c}{0.096} & 0.092  & \multicolumn{1}{c}{0.116 } & \multicolumn{1}{c}{0.159} & 0.220  & \multicolumn{1}{c}{0.132} & \multicolumn{1}{c}{0.207} & 0.160  \\
			\midrule
			\multirow{3}[2]{*}{VOS~\cite{li2017benchmark}} & maxF  & \textcolor[rgb]{1.000, 0.000, 0.000}{\textbf{0.767}} & \multicolumn{1}{c}{\textcolor[rgb]{0.000, 0.690, 0.314}{\textbf{0.747}}} & 0.649  & \multicolumn{1}{c}{\textcolor[rgb]{0.000, 0.439, 0.753}{\textbf{0.742}}} & \multicolumn{1}{c}{0.735} & \multicolumn{1}{c}{0.724} & 0.735  & \multicolumn{1}{c}{0.742} & \multicolumn{1}{c}{0.670} & 0.690  & \multicolumn{1}{c}{0.546} & \multicolumn{1}{c}{0.426} & 0.526  \\
			& S-M   & \textcolor[rgb]{1.000, 0.000, 0.000}{\textbf{0.831}} & \multicolumn{1}{c}{\textcolor[rgb]{0.000, 0.690, 0.314}{\textbf{0.827}}} & 0.695  & \multicolumn{1}{c}{\textcolor[rgb]{0.000, 0.439, 0.753}{\textbf{0.819}}} & \multicolumn{1}{c}{0.792} & \multicolumn{1}{c}{0.798} & 0.818  & \multicolumn{1}{c}{0.818} & \multicolumn{1}{c}{0.742} & 0.712  & \multicolumn{1}{c}{0.624} & \multicolumn{1}{c}{0.557} & 0.576  \\
			& MAE   & \textcolor[rgb]{0.000, 0.690, 0.314}{\textbf{0.066 }} & \multicolumn{1}{c}{\textcolor[rgb]{1.000, 0.000, 0.000}{\textbf{0.065}}} & 0.115  & \multicolumn{1}{c}{0.073} & \multicolumn{1}{c}{0.075} & \multicolumn{1}{c}{\textcolor[rgb]{1.000, 0.000, 0.000}{\textbf{0.065}}} & \textcolor[rgb]{0.000, 0.439, 0.753}{\textbf{0.068}} & \multicolumn{1}{c}{0.078} & \multicolumn{1}{c}{0.099} & 0.162  & \multicolumn{1}{c}{0.145} & \multicolumn{1}{c}{0.236} & 0.163  \\
			\bottomrule
		\end{tabular}%
	}
	\label{tab:result_end}%
\end{table*}%

So, due to the above mentioned issues, we propose a novel filtering scheme, aiming to exclude the less-trustworthy or redundant training instances, see below.\\
\underline{\textbf{1)}} For each frame in the new training set, we measure the consistency degree (we choose the S-Measure, but not limited to it) between its motion saliency map and the VSOD result produced by the target SOTA method.\\
\underline{\textbf{2)}} For each T frames in the new training set, only one frame with the largest consistency degree---this consistency degree is usually positively correlated to the trustworthy degree regarding the VSOD predictions made by the target SOTA method, will be remained (see the detailed ablation study on T in Table~\ref{tab:T}).

\subsubsection{New Round Of Network Training}
Once the new training set has been constructed, we will conduct a new round of network training on it.
However, we can not directly retrain the target SOTA model using this new training set, because it only consists of individual video frames without any temporal information; i.e., our new training set only preserves spatial information, while the SOTA models need to be fed by both spatial and temporal information.
So, we choose to set up a completely new model with an identical network structure to the localization branch demonstrated in Fig.~\ref{fig:Net}, and this new model will be trained over this new training set by using the common thread supervised training protocol (Eq.~\ref{eq:BCE}), and its output will be our final VSOD results with much improved performance compared with the target SOTA method.

Specifically, though this new round of training requires additional time cost, the performance gain can still benefit scenarios without speed requirements.

\section{Experiments}
\subsection{Datasets}
We have evaluated our method on five widely used public available datasets, including Davis~\cite{perazzi2016benchmark}, Segtrack-v2~\cite{li2013video}, Visal~\cite{wang2015consistent}, DAVSOD~\cite{fan2019shifting_ssav}, and VOS~\cite{li2017benchmark}.
\begin{itemize}
	\item
	Davis dataset contains 50 video sequences with 3455 frames in total, and most of its sequences only contain moderate motions.
	\item
	Segtrack-v2 dataset contains 13 video sequences (exclude the penguin sequence) with 1024 frames in total, containing complex backgrounds and variable motion patterns, which is more challenging than the Davis dataset generally.
	\item
	Visal dataset contains 17 video sequences with 963 frames in total, and this dataset is a relatively simple one than others.
	\item
	DAVSOD dataset contains 226 video sequences with 23938 frames in total, which is the most challenging dataset in the field, involving various object instances, different motion patterns, and saliency shifting between different objects.
	\item
	VOS dataset contains 40 video sequences with 24177 frames in total, yet only 1540 frames were annotated well, in which the sequences are all obtained in indoor scenes.
\end{itemize}

\subsection{Implementation Details}
We have implemented our method on a PC with an Intel(R) Xeon(R) CPU, Nvidia GTX2080Ti GPU (with 11G RAM) and 64G RAM.
We use the DAVIS-TR~\cite{perazzi2016benchmark} as the initial training set to train our motion quality perception model (MQPM).
Also, an ADAM optimizer~\cite{kingma2014adam} is applied to update the network parameters.
We set the batch size to 8 which takes almost all GPU memory.
The initial learning rate is set to 10e-3.
To avoid over-fitting problem, we have adopted the random horizontal flips for data augmentation.

\begin{figure*}[http]
	\centering
	\includegraphics[width=\linewidth]{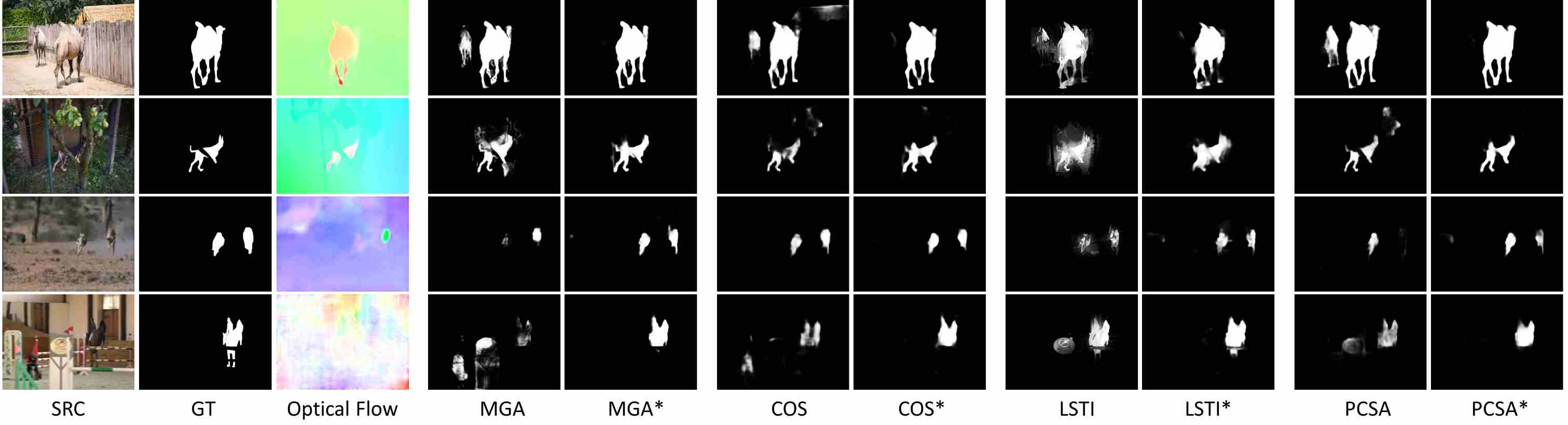}
	\caption{Qualitative comparisons between several most representative target SOTA methods and the corresponding VSOD results after using our novel learning scheme.}
	\label{fig:MM}
\end{figure*}

\begin{table*}[t]
	\centering
	\caption{Quantitative comparisons of several most representative SOTA methods (SSAV19, MGA19, COS19, LSTI20, and PCSA20) vs. their improved results by using our novel learning scheme.}
	\resizebox{0.8\textwidth}{!}{
		\begin{tabular}{|r|c|cc|cc|cc|cc|cc|}
			\toprule
			Dataset & Metric & SSAV\cite{fan2019shifting_ssav} & SSAV* & MGA\cite{li2019motion_mga} & MGA*  & COS\cite{lu2019see} & COS*  & LSTI\cite{chen2019improved} & LSTI* & PCSA\cite{gupyramid_pcsa} & PCSA* \\
			\midrule
			\multirow{3}[2]{*}{Davis~\cite{perazzi2016benchmark}} & maxF  & 0.861  & \textcolor[rgb]{0.000, 0.690, 0.314}{\textbf{0.894}} & \textcolor[rgb]{0.000, 0.439, 0.753}{\textbf{0.892}} & \textcolor[rgb]{1.000, 0.000, 0.000}{\textbf{0.900}} & 0.875  & \textcolor[rgb]{0.000, 0.439, 0.753}{\textbf{0.892}} & 0.850  & 0.863  & 0.880  & \textcolor[rgb]{0.000, 0.690, 0.314}{\textbf{0.894}} \\
			& S-M   & 0.893  & 0.906  & \textcolor[rgb]{0.000, 0.690, 0.314}{\textbf{0.910}} & \textcolor[rgb]{1.000, 0.000, 0.000}{\textbf{0.914}} & 0.902  & 0.909  & 0.876  & 0.889  & 0.902  & \textcolor[rgb]{0.000, 0.439, 0.753}{\textbf{0.909}} \\
			& MAE   & 0.023  & 0.020  & 0.023  & \textcolor[rgb]{0.000, 0.690, 0.314}{\textbf{0.018}} & 0.020  & \textcolor[rgb]{1.000, 0.000, 0.000}{\textbf{0.017}} & 0.034  & 0.024  & 0.022  & \textcolor[rgb]{0.000, 0.439, 0.753}{\textbf{0.019}} \\
			\midrule
			\multirow{3}[2]{*}{SegV2\cite{li2013video}} & maxF  & 0.801  & \textcolor[rgb]{0.000, 0.439, 0.753}{\textbf{0.836}} & 0.821  & 0.835  & 0.801  & 0.815  & \textcolor[rgb]{0.000, 0.690, 0.314}{\textbf{0.858}} & \textcolor[rgb]{1.000, 0.000, 0.000}{\textbf{0.862}} & 0.810  & 0.835  \\
			& S-M   & 0.851  & \textcolor[rgb]{0.000, 0.439, 0.753}{\textbf{0.880}} & 0.865  & \textcolor[rgb]{0.000, 0.690, 0.314}{\textbf{0.882}} & 0.850  & 0.866  & 0.870  & \textcolor[rgb]{1.000, 0.000, 0.000}{\textbf{0.891}} & 0.865  & \textcolor[rgb]{0.000, 0.439, 0.753}{\textbf{0.880}} \\
			& MAE   & 0.023  & \textcolor[rgb]{0.000, 0.439, 0.753}{\textbf{0.019}} & 0.030  & 0.028  & 0.020  & \textcolor[rgb]{0.000, 0.690, 0.314}{\textbf{0.018}} & 0.025  & \textcolor[rgb]{1.000, 0.000, 0.000}{\textbf{0.016}} & 0.025  & 0.020  \\
			\midrule
			\multirow{3}[2]{*}{Visal\cite{wang2015consistent}} & maxF  & 0.939  & 0.935  & 0.933  & 0.933  & \textcolor[rgb]{1.000, 0.000, 0.000}{\textbf{0.966}} & \textcolor[rgb]{0.000, 0.690, 0.314}{\textbf{0.956}} & 0.905  & 0.916  & 0.940  & \textcolor[rgb]{0.000, 0.439, 0.753}{\textbf{0.942}} \\
			& S-M   & 0.943  & 0.940  & 0.936  & 0.931  & \textcolor[rgb]{1.000, 0.000, 0.000}{\textbf{0.965}} & \textcolor[rgb]{0.000, 0.690, 0.314}{\textbf{0.955}} & 0.916  & 0.928  & \textcolor[rgb]{0.000, 0.439, 0.753}{\textbf{0.946}} & \textcolor[rgb]{0.000, 0.439, 0.753}{\textbf{0.946}} \\
			& MAE   & 0.020  & 0.017  & 0.017  & 0.015  & \textcolor[rgb]{0.000, 0.690, 0.314}{\textbf{0.011}} & \textcolor[rgb]{1.000, 0.000, 0.000}{\textbf{0.010}} & 0.033  & 0.022  & 0.017  & \textcolor[rgb]{0.000, 0.439, 0.753}{\textbf{0.014}} \\
			\midrule
			\multirow{3}[2]{*}{DAVSOD\cite{fan2019shifting_ssav}} & maxF  & 0.603  & \textcolor[rgb]{1.000, 0.000, 0.000}{\textbf{0.699}} & 0.640  & \textcolor[rgb]{0.000, 0.439, 0.753}{\textbf{0.672}} & 0.614  & 0.643  & 0.585  & 0.627  & 0.655  & \textcolor[rgb]{0.000, 0.690, 0.314}{\textbf{0.680}} \\
			& S-M   & 0.724  & \textcolor[rgb]{1.000, 0.000, 0.000}{\textbf{0.774}} & 0.738  & \textcolor[rgb]{0.000, 0.690, 0.314}{\textbf{0.755}} & 0.725  & 0.736  & 0.695  & 0.718  & 0.741  & \textcolor[rgb]{0.000, 0.439, 0.753}{\textbf{0.751}} \\
			& MAE   & 0.092  & \textcolor[rgb]{1.000, 0.000, 0.000}{\textbf{0.071}} & 0.084  & \textcolor[rgb]{0.000, 0.690, 0.314}{\textbf{0.075}} & 0.096  & 0.086  & 0.106  & 0.093  & 0.086  & \textcolor[rgb]{0.000, 0.439, 0.753}{\textbf{0.077}} \\
			\midrule
			\multirow{3}[2]{*}{VOS\cite{li2017benchmark}} & maxF  & 0.742  & \textcolor[rgb]{1.000, 0.000, 0.000}{\textbf{0.767}} & 0.735  & \textcolor[rgb]{0.000, 0.439, 0.753}{\textbf{0.755}} & 0.724  & \textcolor[rgb]{0.000, 0.690, 0.314}{\textbf{0.758}} & 0.649  & 0.690  & 0.747  & 0.758  \\
			& S-M   & 0.819  & \textcolor[rgb]{1.000, 0.000, 0.000}{\textbf{0.831}} & 0.792  & 0.811  & 0.798  & 0.810  & 0.695  & 0.722  & \textcolor[rgb]{0.000, 0.690, 0.314}{\textbf{0.827}} & \textcolor[rgb]{0.000, 0.439, 0.753}{\textbf{0.824}} \\
			& MAE   & 0.073  & 0.066  & 0.075  & 0.066  & \textcolor[rgb]{0.000, 0.439, 0.753}{\textbf{0.065}} & \textcolor[rgb]{0.000, 0.690, 0.314}{\textbf{0.063}} & 0.115  & 0.101  & \textcolor[rgb]{0.000, 0.439, 0.753}{\textbf{0.065}} & \textcolor[rgb]{1.000, 0.000, 0.000}{\textbf{0.057}} \\
			\bottomrule
		\end{tabular}%
	}
	\label{tab:other method}%
\end{table*}%

\subsection{Evaluation Metrics}
In order to accurately measure the consistency between the predicted VSOD and the manually annotated ground truth, we adopt three common used evaluation metrics, including the maximum F-measure value (maxF)~\cite{achanta2009frequency}, the mean absolute error (MAE)~\cite{perazzi2012saliency}, and the structure measure value (S-measure)~\cite{fan2017structure}.

\subsection{Component Evaluation}
\label{sec:CE}
\begin{figure}[!h]
	\centering
	\includegraphics[width=1\linewidth]{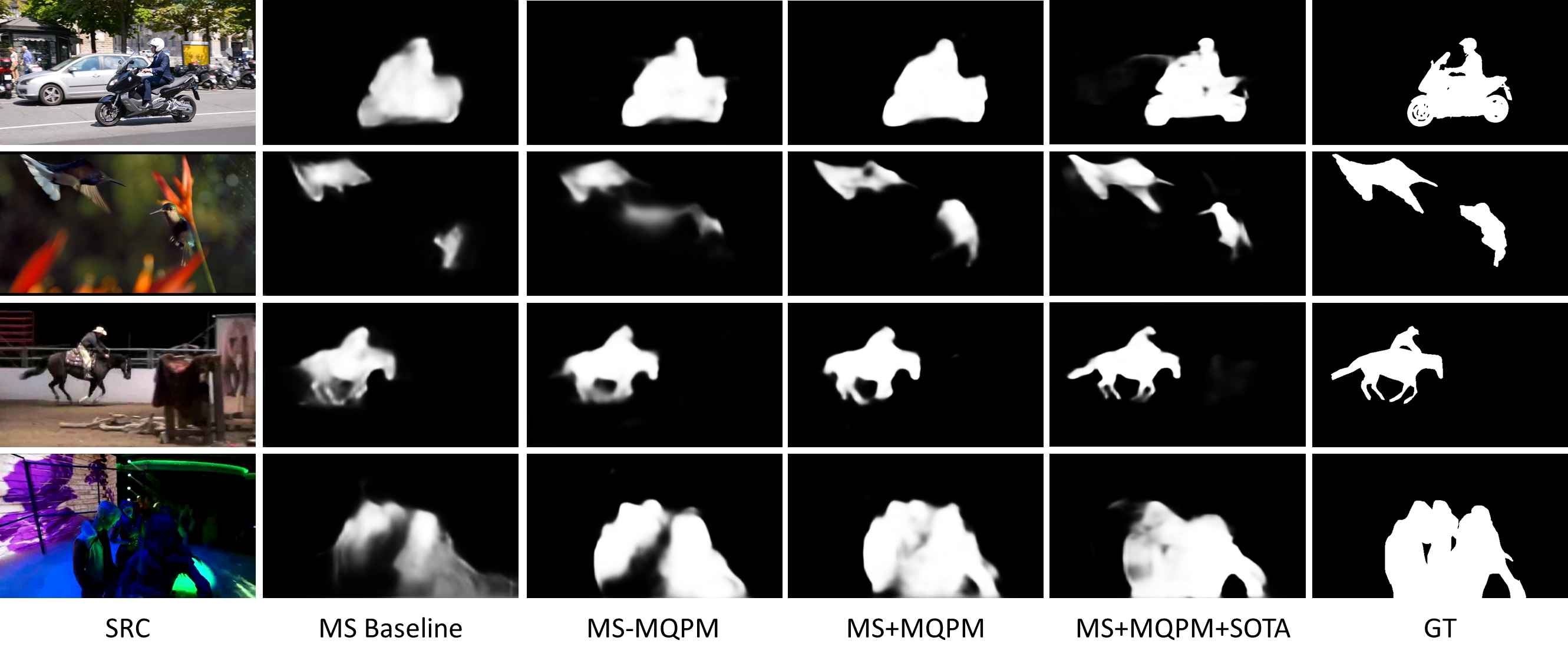}
    \vspace{-0.6cm}
	\caption{The corresponding qualitative demonstrations regarding the component evaluations in Table~\ref{tab:SC}, in which the ``$\rm MS+MQPM+SOTA$'' has achieved the best performance.}
    \vspace{-0.2cm}
	\label{fig:self contrast}
\end{figure}

We have conducted an extensive component evaluation to verify the effectiveness of our proposed motion quality perception module (MQPM), and the quantitative results can be found in Table~\ref{tab:SC}.
Meanwhile, the corresponding qualitative demonstrations regarding this component evaluation can be found in Fig.~\ref{fig:self contrast}.

As is shown in Table~\ref{tab:SC}, the performance of the learned motion saliency, which is denoted by ``MS'' and it can be obtained via $\Theta(\rm OF_\emph{i})$ as mentioned in Eq.~\ref{eq:MQS}, have exhibited the worst performance in all the adopted metrics.
Then, by using our MQPM (Sec.~\ref{sec:MQPM}) to formulate a new training set (MS will be applied as the pseudo-GTs), the overall performance can be improved significantly (denoted by ``MS$+$MQPM''), e.g., the maxF metric value in the VOS dataset has been increased from \underline{40.5}\% to \underline{62.7}\%.
Notice that we can not achieve such performance improvements via the randomly assembled key frames from the training set, and we denote such implementation as ``MS$-$MQPM'', of which the overall performance is quite similar to the original MS baseline. For example, in the \emph{breakdance} video sequence of the Davis testing set, the MQPM has selected 16 high-quality key frames.
For fair comparison, the ``MS$-$MQPM'' randomly select 16 frames as the key frames.

Since the object boundaries are usually blur in the MS baseline, the overall performance of the above re-trained model (i.e., ``MS$+$MQPM'') is limited.
Thus, we further resort our data filtering strategy (Sec.~\ref{sec:DF}) to introduce the target SOTA results as the high-quality pseudo-GTs, of which the corresponding results are shown in the last row of Table~\ref{tab:SC} with the highest scores in all metrics, showing the effectiveness of our data filtering strategy.

Also, it should be noted that we have simply chosen the SSAV~\cite{fan2019shifting_ssav} as the target SOTA method here, because the off-the-shelf SSAV model was pre-trained using the identical training set as our method, which can avoid the data leakage problem.

\subsection{Ablation Study}
\label{sec:AS}
As we have mentioned in Sec.~\ref{sec:DF}, there are almost 30\% video frames in the original testing set which will be predicted to contain high-quality motions (we abbreviate it as the high-quality frames).
Due to the reasons we have mentioned in Sec.~\ref{sec:DF}, we believe that it is time-consuming and not necessary to use all these high-quality frames to start a new round of training.
Thus, the main purpose of our data filtering strategy is to automatically keep a small subgroup of high-quality frames as the final training set.

Thus far, we have conducted an extensive ablation study regarding the parameter T, and the detailed results can be found in Table~\ref{tab:T}.
We choose $\rm T=\{1/10, 1/5, 1/4, 1/3, 1/2, 1\}$ respectively, in which $\rm T=1$ means to use all those high-quality video frames as the new training set, and $\rm T=1/5$ denotes only one frame with the largest consistency degree will be remained for each 5 consecutive high-quality frames.
As is shown in Table~\ref{tab:T}, the overall performance of our method is moderately sensitive to the choice of T, in which the overall performance via $\rm T=1/5$ have exhibited the best performance in general, and a clear performance degradation can be found when we assign $\rm T=1/10$.
So, we set $\rm T=1/5$ as the optimal choice to strike the trade-off between performance and efficiency.

\begin{table*}[htbp]
	\centering
	\caption{Runtime comparisons, where we have excluded the training time (i.e., the FPS provided here is only the inference speed), because the training procedure may only need to be conducted only once for many video saliency based subsequent applications. Also, our method takes about 80s to construct the new training set, and another 600s to conduct the fine-tuning in 5 epoches (this will vary with the training set size); for a single testing frame, it takes about 0.03s to inference SOD result.}
	\vspace{-0.1cm}
	\resizebox{0.9\textwidth}{!}{
		\begin{tabular}{|c|cccccccccc|c}
			\toprule
			Methods & Ours & PCSA20~\cite{gupyramid_pcsa} & LSTI20~\cite{chen2019improved} & SSAV19~\cite{fan2019shifting_ssav} & MGA19~\cite{li2019motion_mga} & COS19~\cite{lu2019see} & PDBM18~\cite{song2018pyramid_pdbm} & SCOM18~\cite{chen2018scom_scom} & SFLR17~\cite{chen2017video} & SGSP17~\cite{liu2016saliency}\\
			\midrule
			FPS   & 33 & \textbf{110} & 0.7   & 20.0  & 14.0  & 0.4   & 20.0  & 0.03  & 0.3 & 0.1 \\
			Platform & GTX2080Ti & GTXTitanXp &  GTX1080Ti & GTXTianX & GTX2080Ti & GTX2080Ti & GTXTitanX & GTXTitanX & GTX970 & CPU\\
			\bottomrule
		\end{tabular}%
	}
	\label{tab:FPS}%
\end{table*}%

\subsection{Comparisons to the SOTA methods}
We have compared our method with 12 most representative SOTA methods,
including PCSA20~\cite{gupyramid_pcsa},LSTI20~\cite{chen2019improved}, SSAV19~\cite{fan2019shifting_ssav}, MGA19~\cite{li2019motion_mga}, COS19~\cite{lu2019see}, CPD19~\cite{wu2019cascaded_cpd}, PDBM18~\cite{song2018pyramid_pdbm}, MBNM18~\cite{li2018unsupervised}, SFLR17~\cite{chen2017video},
SGSP17~\cite{liu2016saliency}, STBP17~\cite{xi2016salient} and SCOM18~\cite{chen2018scom_scom}.


As is shown in Table~\ref{tab:result_end}, all quantitative results have indicated that our method (we take the SSAV as the target SOTA model here) have significantly outperformed these compared SOTA methods for all tested datasets excepting the Visal dataset, showing the performance superiority of our method.
In fact, the Visal dataset may be a bit different to other datasets, i.e., the Visal dataset is dominated by color information, in which the motion clues are usually at the second place to determine the true saliency.
As a result, the COS19, which is heavily rely on the spatial domain, has exhibited the best performance in the Visal dataset.
Also, we have provided the qualitative comparisons in Fig.~\ref{fig:result end vision}, where our VSOD results are more consistent to the GT than those compared SOTA methods.

Moreover, our method can be applied to any other SOTA VSOD methods to get its performance further improved.
To show such advantage, we have provided the direct comparisons between several most representative SOTA methods and their improved versions after using our learning scheme.
As is shown in Table~\ref{tab:other method}, our method can make averagely \underline{5}\% performance improvement generally and almost \underline{9.6}\% regarding the best case (maxF), and the corresponding qualitative comparisons can be found in Fig.~\ref{fig:MM}.

Also, we have conducted the running time comparisons to the SOTA methods in Table~\ref{tab:FPS}, in which our method has achieved the real-time speed with 33 FPS during the inference phase. Although our total time is a bit time-consuming, there are still advantages compared to other methods.

\section{Conclusion}
In this paper, we have proposed a universal scheme to boost the SOTA methods within a semi-supervised manner.
The key components in our method include:
\underline{\textbf{1)}} The motion quality perception module, which was used to select a subgroup of high-quality frames from the original testing set to construct a new training set;
\underline{\textbf{2)}} Data filtering scheme, which was used as a double-check to ensure the overall quality of the newly constructed training set.
We have conducted an extensive quantitative evaluation to respectively show the effectiveness regarding these two components.


\vspace{-0.1cm}
\bibliographystyle{IEEEtran}
\bibliography{TIP_reference}

\end{document}